\documentclass[11pt,a4paper]{article}
\usepackage[hyperref]{stylesheets/acl2020/acl2020}
\usepackage{times}
\usepackage{latexsym}

\usepackage{microtype}

\aclfinalcopy 
\def\aclpaperid{3180} 

\usepackage[utf8]{inputenc} 
\usepackage{lipsum}
\usepackage{todonotes}
\usepackage{enumitem}
\usepackage{tikz}
\usetikzlibrary{positioning}
\usepackage{mathtools}
\usepackage{multirow}
\usepackage{amsmath}
\usepackage{amssymb}
\usepackage{booktabs}
\usepackage{rotating}
\usepackage{xcolor}
\newcommand{\rt}[1]{\begin{rotate}{30}#1\end{rotate}}

\title{Learning and Evaluating Emotion Lexicons for 91 Languages}

\author{ Sven Buechel, Susanna Rücker, and Udo Hahn\vspace*{3pt}\\
         \tt{\{sven.buechel|susanna.ruecker|udo.hahn\}@uni-jena.de}\vspace*{3pt}\\
         Jena University Language and Information Engineering (JULIE) Lab \\
         Friedrich-Schiller-Universit\"at Jena, Jena, Germany \\
         \texttt{https://julielab.de}
         }

\date{}

\begin{document}
\maketitle
\begin{abstract}
Emotion lexicons describe the affective meaning of words and thus constitute a centerpiece for advanced sentiment and emotion analysis. Yet, manually curated lexicons are only available for a handful of languages, leaving most languages of the world without such a precious resource for downstream applications. Even worse, their coverage is often limited both in terms of the lexical units they contain and the emotional variables they feature. In order to break this bottleneck, we here introduce a methodology for creating almost arbitrarily large emotion lexicons for any target language. Our approach requires nothing but a source language emotion lexicon, a bilingual word translation model, and a target language embedding model. Fulfilling these requirements for 91 languages, we are able to generate representationally rich high-coverage lexicons comprising eight emotional variables with more than 100k lexical entries each. We evaluated the automatically generated lexicons against human judgment from 26 datasets, spanning 12 typologically diverse languages, and found that our approach produces results in line with state-of-the-art \textit{monolingual} approaches to lexicon creation and even \textit{surpasses human reliability} for some languages and variables. Code and data are available at \href{https://github.com/JULIELab/MEmoLon}{\tt \small github.com/JULIELab/MEmoLon} archived under DOI \href{https://doi.org/10.5281/zenodo.3779901}{\tt \small 10.5281/zenodo.3779901}.
\end{abstract}

\section{Introduction}

An emotion lexicon is a lexical repository which encodes the affective meaning of individual words (lexical entries). Most simply, affective meaning can be encoded in terms of \textit{polarity}, i.e., the distinction whether an item is considered as positive, negative, or neutral. This is the case for many well-known resources such as \textsc{WordNet-Affect} \citep{Strapparava04}, \textsc{SentiWordNet} \citep{Baccianella10}, or \textsc{Vader} \citep{Hutton14icws}. Yet, an increasing number of researchers focus on more expressive encodings for affective states inspired by distinct lines of work in psychology \citep{Yu16naacl,Buechel17eacl,Sedoc17eacl,Abdul17acl,Bostan18coling,Mohammad18acl,Troiano19acl}. 

Psychologists, on the one hand, value such lexicons as a controlled set of stimuli for designing experiments, e.g., to investigate patterns of lexical access or the structure of memory \citep{hofmann_affective_2009,monnier_semantic_2008}. NLP researchers, on the other hand, use them to augment the emotional loading of word embeddings \citep{Yu17emnlp,Khosla18}, as additional input to sentence-level emotion models so that the performance of even the most sophisticated neural network gets boosted \citep{Mohammad17wassa,Mohammad18semeval,DeBruyne19arxiv}, or rely on them in a keyword-spotting approach when no training data is available, e.g., for studies dealing with historical language stages \citep{Buechel16lt4dh}.

As with any kind of manually curated resource, the availability of emotion lexicons is heavily restricted to only a few languages whose exact number varies depending on the variables under scrutiny. For example, we are aware of lexicons for 15 languages that encode the emotional variables of Valence, Arousal, and Dominance (see Section \ref{sec:related.emotion}). This number leaves the majority of the world's (less-resourced) languages without such a dataset. In case such a lexicon exists for a particular language, it is often severely limited in size, sometimes only comprising some hundreds of entries \cite{Davidson14}. Yet, even the largest lexicons typically cover only some ten thousands of words, still leaving out major portions of the emotion-carrying vocabulary. This is especially true for languages with complex morphology or productive compounding, such as Finnish, Turkish, Czech, or German. Finally, the diversity of emotion representation schemes adds another layer of complexity. While psychologists and NLP researchers alike find that different sets of emotional variables are complementary to each other \cite{Stevenson07,Pinheiro17,Barnes19nodalida,DeBruyne19arxiv}, \textit{manually} creating emotion lexicons for every language and every emotion representation scheme is virtually impossible.

We here propose an approach based on cross-lingual distant supervision to generate almost arbitrarily large emotion lexicons for any target language and emotional variable, provided the following requirements are met: a source language emotion lexicon covering the desired variables, a bilingual word translation model, and a target language embedding model. By fulfilling these preconditions, we can \textit{automatically} generate emotion lexicons for 91 languages covering ratings for eight emotional variables and hundreds of thousands of lexical entries each. Our experiments reveal that our method is on a par with state-of-the-art monolingual approaches and compares favorably with (sometimes even outperforms) human reliability.

\section{Related Work}

\paragraph{Representing Emotion.}
\label{sec:related.emotion}

Whereas research in NLP has focused for a very long time almost exclusively on \textit{polarity}, more recently, there has been a growing interest in more informative representation structures for affective states by including different groups of emotional variables \cite{Bostan18coling}.
Borrowing from distinct schools of thought in psychology, these variables can typically be subdivided into \textit{dimensional} vs.\ \textit{discrete} approaches to emotion representation \cite{Calvo13}.
The \textit{dimensional} approach assumes that emotional states can be \textit{composed} out of several foundational factors, most noticeably \textit{Valence} (corresponding to polarity), \textit{Arousal} (measuring calmness vs.\ excitement), and \textit{Dominance} (the perceived degree of control in a social situation); VAD, for short \cite{Bradley94}.
Conversely, the \textit{discrete} approach assumes that emotional states can be \textit{reduced} to a small, evolutionary motivated set of basic emotions \cite{Ekman92}. Although the exact division of the set has been subject of hot debates, recently constructed datasets (see Section \ref{sec:data}) most often cover the categories of  \textit{Joy}, \textit{Anger}, \textit{Sadness}, \textit{Fear}, and \textit{Disgust}; BE5, for short.
Plutchik's Wheel of Emotion takes a middle ground between those two positions by postulating emotional categories which are yet grouped into opposite pairs along different levels of intensity \citep{Plutchik80}.

Another dividing line between representational approaches is whether target variables are encoded in terms of (strict) class-membership or scores for numerical strength. In the first case, emotion analysis translates into a (multi-class) classification problem, whereas the latter turns it into a regression problem \citep{Buechel16ecai}.
While our proposed methodology is agnostic towards the chosen emotion format, we will focus on the VAD and BE5 formats here, using numerical ratings (see the examples in Table \ref{tab:examples}) due to the widespread availability of such data. Accordingly, this paper treats word emotion prediction as a regression problem.

\begin{table}[h!]
	\small
	\centering
	\setlength{\tabcolsep}{5pt} 
	\begin{tabular}{l|lll|lllll}
		 & \rt{\textbf{Val}} & \rt{\textbf{Aro}} & \rt{\textbf{Dom}} & \rt{\textbf{Joy}} & \rt{\textbf{Ang}} & \rt{\textbf{Sad}} & \rt{\textbf{Fea}} & \rt{\textbf{Dis}}\\
    \toprule
		\textit{sunshine} 	& 8.1 & 5.3 & 5.4 & 4.2 & 1.2 & 1.3 & 1.3 & 1.2\\
		\textit{terrorism}	& 1.6 & 7.4 & 2.7 & 1.2 & 2.9 & 3.3 & 3.9 & 2.5\\
		\textit{nuclear} 	& 4.3 & 7.3 & 4.1 & 1.4 & 2.2 & 1.9 & 3.2 & 1.6\\
		\textit{ownership} & 5.9 & 4.4 & 7.5 & 2.1 & 1.4 & 1.2 & 1.4 & 1.3 \\
		\bottomrule
	\end{tabular}
	\caption{Sample entries from our English source lexicon described via eight emotional variables: \textbf{Val}ence, \textbf{Aro}usal, \textbf{Dom}inance [VAD], and \textbf{Joy}, \textbf{Ang}er, \textbf{Sad}ness, \textbf{Fea}r, and \textbf{Dis}gust [BE5]. VAD uses 1-to-9 scales (``5'' encodes the neutral value) and BE5 1-to-5 scales (``1'' encodes the neutral value).  
		\label{tab:examples}}
	\vspace{-10pt}
\end{table}

\paragraph{Building Emotion Lexicons.}
\label{sec:related.lexica}

Usually, the ground truth for affective word ratings (i.e., the assignment of emotional values to a lexical item) is acquired in a questionnaire study design where subjects (annotators) receive lists of words which they rate according to different emotion variables or categories. Aggregating individual ratings of multiple annotators then results in the final emotion lexicon \citep{Bradley99anew}. Recently, this workflow has often been enhanced by crowdsourcing \citep{Mohammad13} and best-worst scaling \citep{Kiritchenko16}. 

As a viable alternative to manual acquisition, such lexicons can also be created by automatic means \citep{Bestgen08,Koeper16,Shaikh16}, i.e., by learning to predict emotion labels for unseen words. Researchers have worked on this prediction problem for quite a long time. Early work tended to focus on word statistics, often in combination with linguistic rules \citep{Hatzivassiloglou97acl,Turney03}. More recent approaches focus heavily on word embeddings, either using semi-supervised graph-based approaches \citep{Wang16words,Hamilton16emnlp,Sedoc17eacl} or  fully supervised methods \citep{Rosenthal15,Li17,Rothe16,Du16}. Most important for this work, \citet{Buechel18naacl} report on near-human performance using a combination of \textsc{fastText} vectors and a multi-task feed-forward network (see Section \ref{sec:setup}). While this line of work can add new words, it does not extend lexicons to other emotional variables or languages.

A relatively new way of generating novel labels is \textit{emotion representation mapping} (ERM), an annotation projection that translates ratings from one emotion format into another, e.g., mapping VAD labels into BE5, or vice versa \citep{Hoffmann12,Buechel16ecai,Buechel18coling,Alarcao17,Landowska18mapping,Zhou18arxiv,Park19arxiv}. While our work uses ERM to add additional emotion variables to the source lexicon, ERM alone can neither increase the coverage of a lexicon, nor adapt it to another language.

\paragraph{Translating Emotions.}
\label{sec:related.translation}

The approach we propose is strongly tied to the observation by \citet{Leveau12} and \citet{Warriner13} who found---comparing a large number of existing emotion lexicons of different languages---that translational equivalents of words show strong stability and adherence to their emotional value. Yet, their work is purely descriptive. They do not exploit their observation to create new ratings, and only consider manual rather than automatic translation.

Making indirect use of this observation, \citet{Mohammad13} offer machine-translated versions of their \textit{NRC Emotion Lexicon}. Also, many approaches in cross-lingual sentiment analysis (on the sentence-level) rely on  translating polarity lexicons \citep{Abdalla17ijcnlp,Barnes18acl}. Perhaps most similar to our work, \citet{Chen14acl} create (polarity-only) lexicons for 136 languages by building a multilingual word graph and propagating sentiment labels through that graph. Yet, their method is restricted to high frequency words---their lexicons cover between 12 and 4,653 entries, whereas our approach exceeds this limit by more than two orders of magnitude.

Our methodology also resembles previous work which models word emotion for historical language stages \citep{Cook10,Hamilton16emnlp,Hellrich18coling,Li19brm}. Work in this direction typically comes up with a set of seed words with assumingly \textit{temporally stable} affective meaning (our work assumes stability against translation) and then uses distributional methods to derive emotion ratings in the target language stage. However, gold data for the target language (stage) is usually inaccessible, often preventing evaluation against human judgment. 
In contrast, we here propose several alternative evaluation set-ups as an integral part of our methodology.

\begin{figure*}[t!]
\vspace*{-3mm}
    \centering
    \vspace*{-3mm}
    \includegraphics[width=\textwidth]{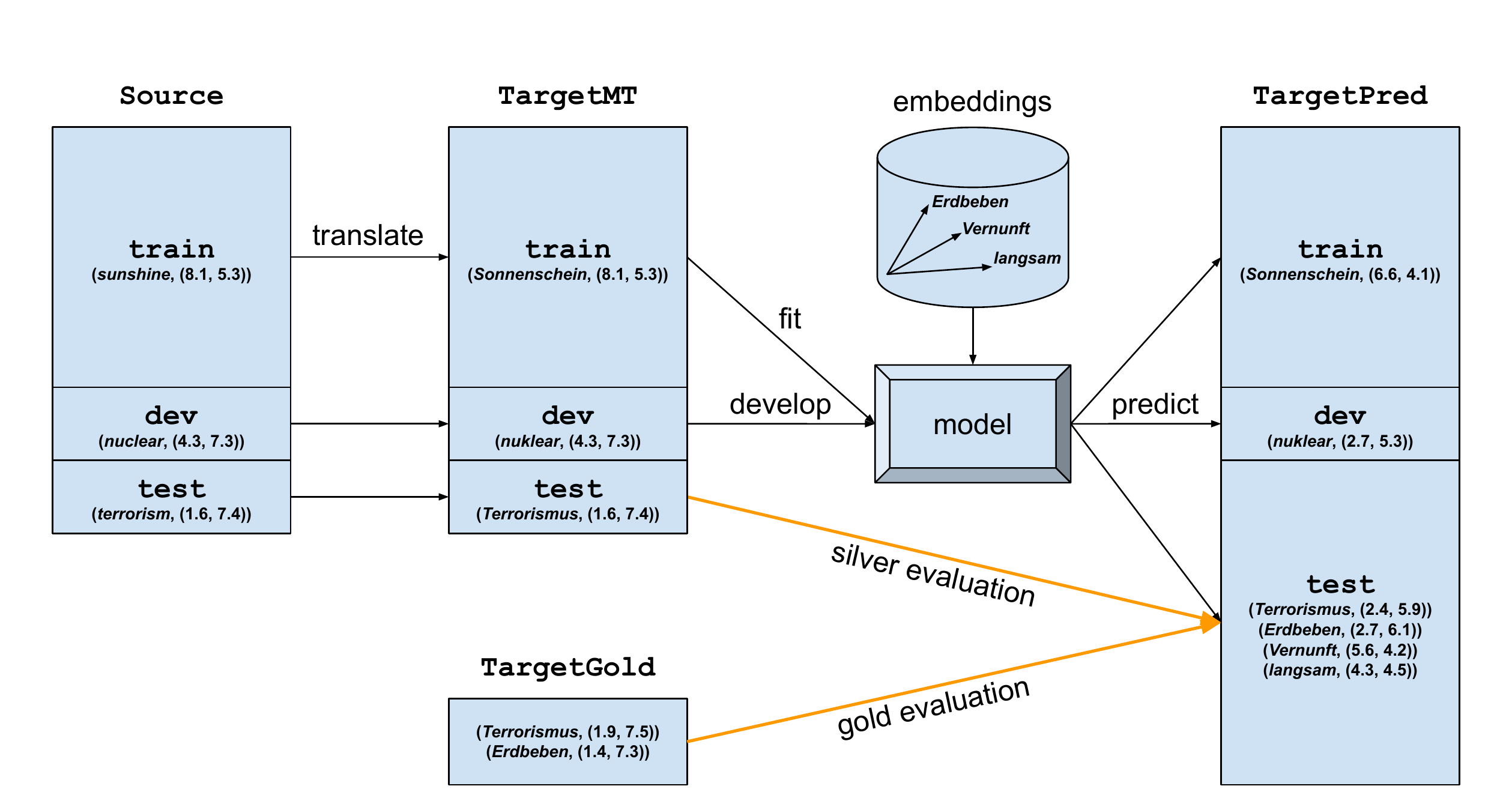}
    \caption{\label{fig:methodology}
    Schematic view on the methodology for generating and evaluating an emotion lexicon for a given target language based on source language supervision. Included is a toy example starting with an English VA lexicon (\textit{sunshine, nuclear, terrorism} and the associated numerical scores for Valence and Arousal) and resulting in an extended German lexicon which incorporates translated entries with altered VA scores and additional entries originating from the embedding model with newly learned scores.
    }
    \vspace*{-2mm}
    
\end{figure*}

\section{A Novel Approach to  Lexicon Creation}
\label{sec:methodology}

Our methodology integrates (1) cross-lingual generation and expansion of emotion lexicons and (2) their evaluation against gold and silver standard data. Consequently, a key aspect of our workflow design is how data is split into train, dev, and test sets at different points of the generation process. Figure \ref{fig:methodology} gives an overview of our framework including a toy example for illustration.

\paragraph{Lexicon Generation.}
\label{sec:generation_method}

We start with a lexicon (\texttt{Source}) of arbitrary size, emotion format\footnote{This encompasses not only VA(D) and BE5, but also any sort of (real-valued) polarity encodings.}  
and source language which is partitioned into train, dev, and test splits denoted by \texttt{Source-train}, \texttt{Source-dev}, and \texttt{Source-test}, respectively. Next, we leverage a bilingual word translation model between source and desired target language to build the first target-side emotion lexicon denoted as \texttt{TargetMT}.  Source words are translated according to the model, whereas target-side emotion labels are simply copied from the source to the target (see Section \ref{sec:related.translation}). Entries are assigned to train, dev, or test set according to their source-side assignment (cf.\  Figure \ref{fig:methodology}). The choice of our translation service (see below) ensures that each source word receives exactly one translation.

\texttt{TargetMT} is then used as the distant supervisor to train a model that predicts word emotions based on target-side word embeddings. \texttt{TargetMT-train} and \texttt{TargetMT-dev} are used to fit model parameters and optimize hyperparameters, respectively, whereas \texttt{TargetMT-test} is held out for later evaluation. 
Once finalized, the model is used to predict \textit{new labels} for the words in \texttt{TargetMT}, resulting in a second target-side emotion lexicon denoted \texttt{TargetPred}. Our rationale for doing so is that a reasonably trained model should generalize well over the entire \texttt{TargetMT} lexicon because it has access to the target-side embedding vectors. Hence, it may mitigate some of the errors which were introduced in previous steps, either by machine translation or by assuming that source- and target-side emotion are always identical. We validate this assumption in Section \ref{sec:analysis}. We also predict ratings for \textit{all} the words in the embedding model, leading to a large number of new entries.  

The splits are defined as follows: let $MT_\mathrm{train}$, $MT_\mathrm{dev}$, and $MT_\mathrm{test}$ denote the set of words in train, dev, and test split of \texttt{TargetMT}, respectively. Likewise, let $P_\mathrm{train}$, $P_\mathrm{dev}$, and $P_\mathrm{test}$ denote the splits of \texttt{TargetPred} and let $E$ denote the set of words in the embedding model. Then\vspace*{-8pt}

\begin{align*}
    & P_\mathrm{train} &&:= && MT_\mathrm{train}\\
    & P_\mathrm{dev} &&:= && MT_\mathrm{dev} \setminus MT_\mathrm{train} \\
    & P_\mathrm{test} &&:= && (MT_\mathrm{test} \cup E) \setminus (MT_\mathrm{dev} \cup MT_\mathrm{train})
\end{align*}

The above definitions help clarify the way we address polysemy.\footnote{
    In short, our work evades this problem by dealing with lexical entries exclusively on the type- rather than the sense-level. From a lexicological perspective, this may seem like a strong assumption. From a modeling perspective, however, it appears almost obvious as it aligns well with the major components of our methodology, i.e., lexicons, embeddings, and translation. The lexicons we work with follow the design of behavioral experiments: a stimulus (word type) is given to a subject and the response (rating) is recorded. The absence of sense-level annotation simplifies the mapping between lexicon and embedding entries. While  sense embeddings form an active area of research \cite{CamachoCollados18,Chi18}, to the best of our knowledge, type-level embeddings yield state-of-the-art performance in downstream applications. 
}
Ambiguity on the target-side may result in multiple source entries translating to the same target-side word.\footnote{
    Source-side polysemy, in contrast to its target-side counterpart, is less of a problem, because we receive only a single candidate during translation. This may result in cases where the translation misaligns with the copied emotion value in \texttt{TargetMT}. Yet, the prediction step partly mitigates such inconsistencies (see Section \ref{sec:analysis}). 
}
This circumstance leads to ``partial duplicates'' in \texttt{TargetMT}, i.e., groups of entries with the same word type but different emotion values (because they were derived from distinct \texttt{Source} entries). Such overlap could do harm to the integrity of our evaluation since knowledge may ``leak'' from training to validation phase, i.e., by testing the model on words it has already seen during training, although with distinct emotion labels. The proposed data partitioning eliminates such distortion effects. Since partial duplicates receive the same embedding vector, the prediction model assigns the same emotion value to both, thus merging them in \texttt{TargetPred}.

\paragraph{Evaluation Methodology.}
\label{sec:eval_method}

The main advantage of the above generation method is that it allows us to create large-scale emotion lexicons for languages for which gold data is lacking. But if that is the case, how can we assess the quality of the generated lexicons?
Our solution is to propose two different evaluation scenarios---a \textit{gold evaluation} which is a strict comparison against human judgment, meaning that it is limited to languages where such data (denoted \texttt{TargetGold}) is available, and a \textit{silver evaluation} which substitutes human judgments by automatically derived ones (silver standard) which is feasible for any language in our study. The rationale is that if both, gold and silver evaluation, strongly agree with each other, we can use one as proxy for the other when no target-side gold data exists (examined in Section \ref{sec:gold-silver-agreement}).

Note that our lexicon generation approach consists of two major steps, \textit{translation} and \textit{prediction}. However, these two steps are not equally important for each generated entry in \texttt{TargetPred}. Words, such as  German \textit{Sonnenschein} for which a translational equivalent already exists in the \texttt{Source} (``sunshine''; see Figure \ref{fig:methodology}), mainly rely on translation, while the prediction step acts as an optional refinement procedure. In contrast, the prediction step is crucial for words, such as \textit{Erdbeben}, whose translational equivalents (``earthquake'') are missing in the \texttt{Source}. Yet, these words also depend on the translation step for producing training data.

These considerations are important for deciding which words to evaluate on. We may choose to base our evaluation on the full \texttt{TargetPred} lexicon, including words from the training set---after all, the word emotion model does not have access to \textit{any} target-side gold data. The problem with this approach is that it merges words that mainly rely on \textit{translation}, because their equivalents are in the \texttt{Source}, and those which largely depend on \textit{prediction}, because they are taken from the embedding model. In this case, generalizability of evaluation results becomes questionable.

Thus, our evaluation methodology needs to fulfill the following two requirements:  (1) evaluation must not be performed on translational equivalents of the \texttt{Source} entries to which the model already had access during training (e.g., \textit{Sonnenschein} and \textit{nuklear} in our example from Figure \ref{fig:methodology}); but, on the other hand, (2) a reasonable number of instances must be available for evaluation (ideally, as many as possible to increase reliability). The intricate cross-lingual train-dev-test set assignment of our generation methodology is in place so that we meet these two requirements.

In particular, for our silver evaluation, we intersect \texttt{TargetMT-test} with \texttt{TargetPred-test} and compute the correlation of these two sets individually for each emotion variable. Pearson's $r$ will be used as  correlation measure throughout this paper. Establishing a test set at the very start of our workflow, \texttt{Source-test}, assures that there is a relatively large overlap between the two sets and, by extension, that our requirements for the evaluation are met.

The gold evaluation is a somewhat more challenging case, because we can, in general, not guarantee that the overlap of a \texttt{TargetGold} lexicon with \texttt{TargetPred-test} will be of any particular size. For this reason, the words of the embedding model are added to \texttt{TargetPred-test} (see above), maximizing the expected overlap with \texttt{TargetGold}. In practical terms, we intersect \texttt{TargetGold} with \texttt{TargetPred-test} and compute the variable-wise correlation between these sets, in parallel to the silver evaluation.
A complementary strategy for maximizing overlap, by exploiting dependencies between published lexicons, is described below.

\begin{table}[t!b]
    \centering
    \small
    \begin{tabular}{llrl}
\toprule
\textbf{ID} & \textbf{Encoding} &   \textbf{Size} &         \textbf{Citation} \\
\midrule
en1 &       VAD &   1032 &      \scriptsize \citet{Warriner13}            \\
en2 &       VAD &   1034 &      \scriptsize \citet{Bradley99anew}            \\
en3 &     BE5 &   1034 &        \scriptsize \citet{Stevenson07}            \\
es1 &       VAD &   1034 &      \scriptsize \citet{Redondo07}            \\
es2 &        VA &  14031 &      \scriptsize \citet{Stadthagen17}            \\
es3 &        VA &    875 &      \scriptsize \citet{Hinojosa16}            \\
es4 &     BE5 &    875 &        \scriptsize \citet{Hinojosa16}            \\
es5 &     BE5 &  10491 &        \scriptsize \citet{Stadthagen18}            \\
es6 &     BE5 &   2266 &        \scriptsize \citet{Ferre17}            \\
de1 &       VAD &   1003 &      \scriptsize \citet{Schmidtke14}            \\
de2 &        VA &   2902 &      \scriptsize \citet{Vo09}            \\
de3 &        VA &   1000 &      \scriptsize \citet{Kanske10}            \\
de4 &     BE5 &   1958 &        \scriptsize \citet{Briesemeister11}            \\
pl1 &       VAD &   4905 &      \scriptsize \citet{Imbir16}            \\
pl2 &        VA &   2902 &      \scriptsize \citet{Riegel15}            \\
pl3 &     BE5 &   2902 &        \scriptsize \citet{Wierzba15}            \\
zh1 &        VA &   2794 &      \scriptsize \citet{Yu16naacl}            \\
zh2 &        VA &   1100 &      \scriptsize \citet{Yao17}            \\
it  &       VAD &   1121 &      \scriptsize \citet{Montefinese14}            \\
pt  &       VAD &   1034 &      \scriptsize \citet{Soares12}            \\
nl  &        VA &   4299 &      \scriptsize \citet{Moors13}            \\
id  &       VAD &   1487 &      \scriptsize \citet{Sianipar16}            \\
el  &       VAD &   1034 &      \scriptsize \citet{Palogiannidi16lrec}            \\
tr1 &        VA &   2029 &      \scriptsize \citet{Kapucu18}            \\
tr2 &     BE5 &   2029 &        \scriptsize \citet{Kapucu18}            \\
hr  &        VA &   3022 &      \scriptsize \citet{Coso19qjep}            \\
\bottomrule
\end{tabular}
    \caption{Lexicons used for gold evaluation. \textbf{ID}s consist of the respective ISO 639-1 language code plus a cardinal number to distinguish different datasets, if needed; the format of emotion \textbf{Encoding} is specified and \textbf{Size} gives the number of lexical entries per lexicon.
    \vspace*{-10pt}
    } 
    \label{tab:gold_lexica}
\end{table}

\section{Experimental Setup}
\label{sec:setup}

\paragraph{Gold Lexicons and Data Splits.}
\label{sec:data}

We use the English emotion lexicon from \citet{Warriner13} as first part of our \texttt{Source} dataset. This popular resource comprises about 14k entries in VAD format collected via crowdsourcing.
Since manually gathered BE5 ratings are available only for a subset of this lexicon \citep{Stevenson07}, we add BE5 ratings from \citet{Buechel18coling} who used emotion representation mapping (see Section \ref{sec:related.lexica}) to convert the existing VAD ratings, showing that this is about as reliable as human annotation.

As apparent from the previous section, a crucial aspect for applying our methodology is the design of the train-dev-test split of the \texttt{Source} because it directly impacts the amount of words we can test our lexicons on during gold evaluation.
In line with these considerations, we choose the lexical items which are already present in \textsc{Anew} \cite{Bradley99anew} as \texttt{Source-test} set. \textsc{Anew} is the precursor to the version later distributed by \citet{Warriner13}; it is widely used and has been adapted to a wide range of languages. With this choice, it is likely that a resulting \texttt{TargetPred-test} set has a large overlap with the respective \texttt{TargetGold} lexicon.
As for the \texttt{TargetGold} lexicons, we included every VA(D) and BE5 lexicon we could get hold of with more than 500 entries. This resulted in 26 datasets covering 12  quite diverse languages (see Table \ref{tab:gold_lexica}). Note that we also include English lexicons in the gold evaluation. In these cases, no translation will be carried out (\texttt{Source} is identical to \texttt{TargetMT}) so that only the expansion step is validated. Appendix \ref{appendix:data} gives further details on data preparation.

\begin{figure*}[ht]
    \includegraphics[width=\textwidth]{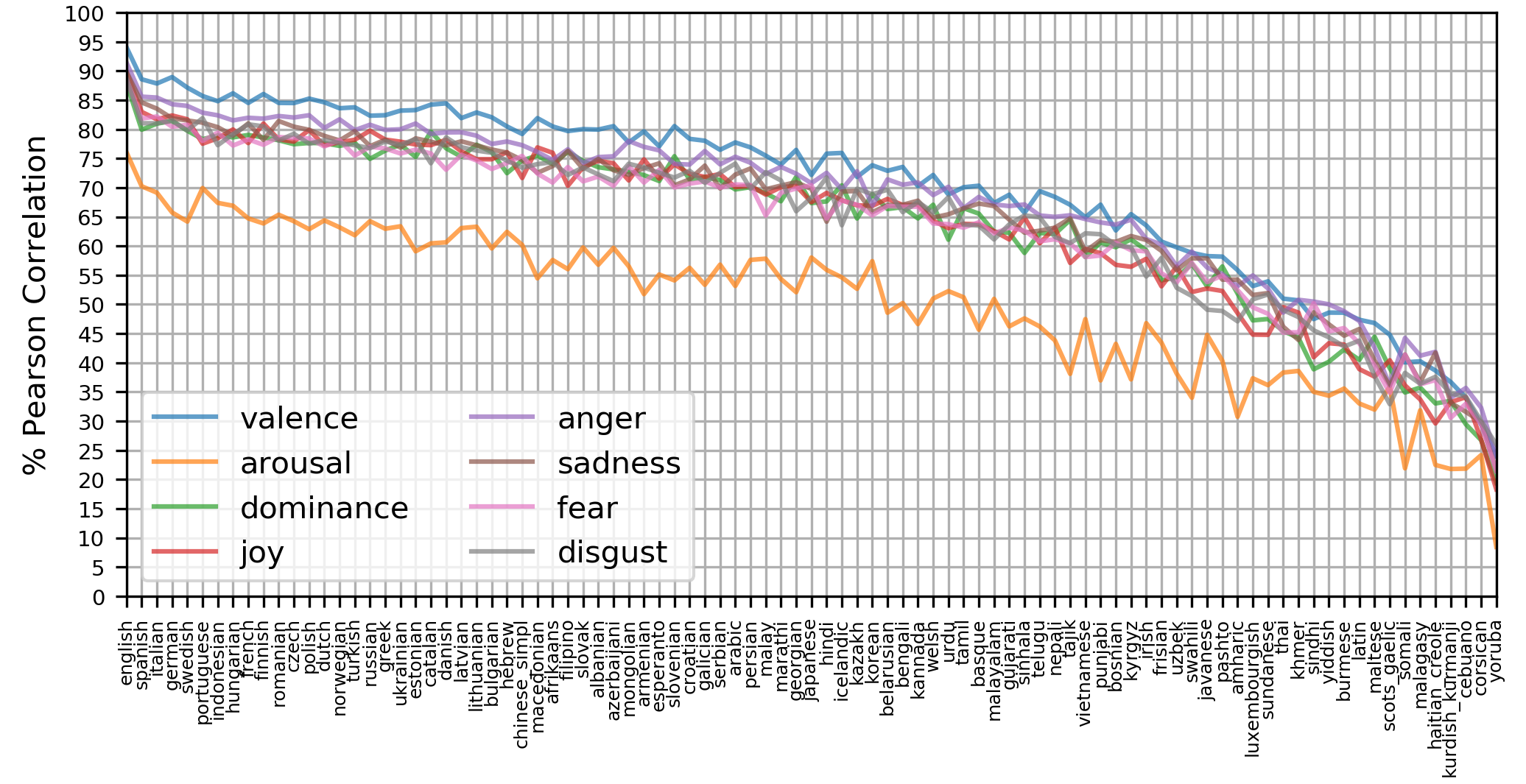}
    \caption{\label{fig:silver}
        Silver evaluation results in Pearson's $r$. Languages (x-axis) are sorted according to mean correlation.
        }
\end{figure*}

\paragraph{Translation.}
\label{sec:results.translation}
We used the \textsc{Google Cloud Trans\-lation} API\footnote{\url{https://cloud.google.com/translate/}} 
to produce word-to-word translation tables. This is a commercial service, total translation costs amount to 160 EUR. API calls were performed in November 2019.

\paragraph{Embeddings.}
We use the \texttt{fastText} embedding models from \newcite{Grave18lrec} trained for 157 languages on the respective \textsc{Wikipedia} and the respective part of \textsc{CommonCrawl}. These resources not only greatly facilitate our work but also increase comparability across languages. The restriction to ``only'' 91 languages comes from intersecting the ones covered by the vectors with the languages covered by the translation service.

\paragraph{Models.}
\label{sec:model}

Since our proposed methodology is agnostic towards the chosen word emotion model, we will re-use models from the literature. In particular, we will rely on the multi-task learning feed-forward network (MTLFFN) worked out by \citet{Buechel18naacl}. This network constitutes the current state of the art for \textit{monolingual} emotion lexicon creation (expanding an existing lexicon for a given language) for many of the datasets in Table \ref{tab:gold_lexica}.

The MTLFFN has two hidden layers of 256 and 128 units, respectively, and takes pre-trained embedding vectors as input. Its distinguishing feature is that hidden layer parameters are shared between the different emotion target variables, thus constituting a mild form of multi-task learning (MTL). We apply MTL to VAD and BE5 variables individually (but not between both groups), thus training two \textit{distinct} emotion models per language, following the outcome of a development experiment. Details are given in Appendix \ref{sec:appendix.model} together with the remainder of the model specifications.

Being aware of the infamous instability of neural approaches \cite{Reimers17emnlp}, we also employ a ridge regression model, an $L_2$ regularized version of linear regression, as a more robust, yet also powerful baseline  \citep{Li17}.

\section{Results}
\label{sec:results}

The size of the resulting lexicons (a complete list is provided in Table \ref{tab:generated-lexica} in the Appendix) ranges from roughly 100k to more than 2M entries mainly depending on the vocabulary of the respective embeddings. 
We want to point out that not every single entry should be considered meaningful because of noise in the embedding vocabulary caused by typos and tokenization errors. However, choosing the ``best'' size for an emotion lexicon necessarily translates into a quality-coverage trade-off for which there is no general solution. Instead, we release the full-size lexicons and leave it to prospective users to apply any sort of filtering they deem appropriate.

\paragraph{Silver Evaluation.}

Figure \ref{fig:silver} displays the results of our silver evaluation. Languages (x-axis) are sorted by their average performance over all variables (not shown in the plot; tabular data given in the Appendix). As can be seen, the evaluation results for English are markedly better than for any other language. This is not surprising since no (potentially error-prone) machine translation was performed.
Apart from that, performance remains relatively stable across most of the languages and starts degrading more quickly only for the last third of them.
In particular, for Valence---typically the easiest variable to predict---we achieve a strong performance of $r > .7$ for 56 languages. On the other hand, for Arousal---typically, the most difficult one to predict---we achieve a solid performance of $r > .5 $ for 55 languages. Dominance and the discrete emotion variables show performance trajectories swinging between these two extremes.
We assume that the main factors for explaining performance differences between languages are the quality of the translation and embedding models which, in turn, both depend on the amount of available text data (parallel or monolingual, respectively).

Comparing MTLFFN and ridge baseline, we find that the neural network reliably outperforms the linear model. 
On average over all languages and variables, the MTL models achieve 6.7\%-points higher Pearson correlation.  Conversely, ridge regression outperforms MTLFFN in only 15 of the total 728 cases (91 languages $\times$ 8 variables).

\begin{table}[t!]
    \centering
    \small
    \vspace*{6pt}
\begin{tabular}{lrr|lll}
\toprule
\textbf{ID} &  \textbf{Shared} & \textbf{(\%)} &  \textbf{Val} &  \textbf{Aro} &  \textbf{Dom} \\
\midrule
en1  &     1032 & 100 &  \textbf{.94} (.87) &  \textbf{.76} (.67) &  \textbf{.88} (.76) \\
en2  &     1034 & 100 &  \textbf{.92} (.92) &  .71 (\textbf{.73}) &  .78 (\textbf{.82}) \\
es1  &      612 & 59 &  \textbf{.91} (.88) &  \textbf{.71} (.70) &  .82 (\textbf{.83}) \\
es2  &     7685 & 54 &  .79 (\textbf{.82}) &  .64 (\textbf{.74}) &  --- \\
es3  &      363 & 41 &  .91 &  .73 &  --- \\
de1  &      677 & 67 &  \textbf{.89} (.87) &  .78 (\textbf{.80}) &  .68 (\textbf{.74}) \\
de2  &     2329 & 80 &  .75 &  .64 &  --- \\
de3  &      916 & 91 &  .80 &  .67 &  --- \\
pl1  &     2271 & 46 &  \textbf{.83} (.74) &  \textbf{.74} (.70) &  .60 (\textbf{.69}) \\
pl2  &     1381 & 47 &  .82 &  .61 &  --- \\
zh1  &     1685 & 60 &  .84 (\textbf{.85}) &  .56 (\textbf{.63}) &  --- \\
zh2  &      701 & 63 &  .84 &  .44 &  --- \\
it   &      660 & 58 &  \textbf{.89} (.86) &  .63 (\textbf{.65}) &  \textbf{.76} (.75) \\
pt   &      645 & 62 &  \textbf{.89} (.86) &  .71 \textbf{(.71)} &  \textbf{.75} (.73) \\
nl   &     2064 & 48 &  \textbf{.85} (.79) &  .58 (\textbf{.74}) &  --- \\
id   &      696 & 46 &  \textbf{.84} (.80) &  \textbf{.64} (.60) &  \textbf{.63} (.58) \\
el   &      633 & 61 &  .86 &  .50 &  .74 \\
tr1  &      721 & 35 &  .75 &  .57 &  --- \\
hr   &     1331 & 44 &  .81 &  .66 &  --- \\
\midrule
\multicolumn{3}{l|}{\textbf{Mn} (all)}      &  \textbf{.85} &  \textbf{.65} &  \textbf{.74} \\
\multicolumn{3}{l|}{\textbf{Mn} (vs. monolingual)}     & \textbf{.87} (.84)   &  .68 (\textbf{.70})  & \textbf{.74} (.74) \\
\bottomrule
\end{tabular}
    \caption{\label{tab:gold-eval-vad}
    Gold evaluation results for VAD (\textbf{Val}ence, \textbf{Aro}usal, \textbf{Dom}inance) in Pearson's $r$. Parentheses give comparative monolingual results from \citet{Buechel18naacl}. \textbf{Shared} words between \texttt{Target\-Gold} and \texttt{TargetPred-test}; \textbf{(\%)}: percentage relative to \texttt{TargetGold}; \textbf{Mn} (all): mean over all datasets; \textbf{Mn} (vs.\ monolingual): mean over datasets with comparative results.
    }

\end{table}

\begin{table}[t!]
\centering
\small
\begin{tabular}{lrr|rrrrr}
\toprule
\textbf{ID} &  \textbf{Shared} & \textbf{(\%)} &  \textbf{Joy} &  \textbf{Ang} &  \textbf{Sad} &  \textbf{Fea} &  \textbf{Dis} \\
\midrule
en3  &    1033 &   99 &  .89 &  .83 &  .80 &  .82 &  .78 \\
es4  &     363 &   41 &  .86 &  .84 &  .84 &  .84 &  .76 \\
es5  &    6096 &   58 &  .64 &  .72 &  .72 &  .72 &  .63 \\
es6  &     992 &   43 &  .80 &  .74 &  .71 &  .72 &  .68 \\
de4  &     848 &   43 &  .80 &  .66 &  .52 &  .68 &  .42 \\
pl3  &    1381 &   47 &  .78 &  .71 &  .66 &  .69 &  .71 \\
tr2  &     721 &   35 &  .77 &  .69 &  .71 &  .70 &  .65 \\
\midrule
\multicolumn{3}{l|}{\textbf{Mean}}  &  \textbf{.79} &  \textbf{.74} &  \textbf{.71} &  \textbf{.74} &  \textbf{.66} \\
\bottomrule
\end{tabular}
\caption{\label{tab:gold-eval-be}
Gold evaluation results for BE5 (\textbf{Joy}, \textbf{Ang}er, \textbf{Sad}ness, \textbf{Fea}r, \textbf{Dis}gust) in Pearson's $r$. \textbf{Shared} words between \texttt{TargetGold} and \texttt{Target\-Pred-test}; \textbf{(\%)}: percentage relative to \texttt{TargetGold}; \textbf{Mean} over all datasets. 
\vspace*{-10pt}}

\end{table}

\paragraph{Gold Evaluation.}

Results for VAD variables on gold data are given in Table \ref{tab:gold-eval-vad}. As can be seen, our lexicons show a good correlation with human judgment and do so robustly, even for less-resourced languages, such as Indonesian (id), Turkish (tr), or Croatian (hr), and across affective variables. Perhaps the strongest negative outliers are the Arousal results for the two Chinese datasets (zh), which are likely to result from the low reliability of the gold ratings (see below).

We compare these results against those from \newcite{Buechel18naacl} which were acquired on the respective \texttt{TargetGold} dataset in a monolingual fashion using 10-fold cross-validation (10-CV). We admit that those results are not fully comparable to those presented here because we use fixed splits rather than 10-CV.
Nevertheless, we find that the results of our cross-lingual set-up are more than competitive, outperforming the monolingual results from \newcite{Buechel18naacl} in 17 out of 30 cases (mainly for Valence and Dominance, less often for Arousal). This is surprising since we use an otherwise identical model and training procedure. 
We conjecture that the large size of the English \texttt{Source} lexicon, compared to most \texttt{TargetGold} lexicons, more than compensates for error-prone machine translation.

Table \ref{tab:gold-eval-be} shows the results for BE5 datasets which are in line with the VAD results. Regarding the ordering of the emotional variables, again, we find Valence to be the easiest one to predict, Arousal the hardest, whereas basic emotions and Dominance take a middle ground.

\paragraph{Comparison against Human Reliability.}
We base this analysis on \textit{inter-study reliability} (ISR), a rather strong criterion for human performance. ISR is computed, per variable,  as the correlation between the ratings from two distinct annotation studies \cite{Warriner13}. Hence, this analysis is restricted to languages where more than one gold lexicon exists per emotion format. We intersect the entries from both gold standards as well as the respective \texttt{TargetPred-test} set and compute the correlation between all three pairs of lexicons. If our lexicon agrees more with one of the gold standards than the two gold standards agree with each other, we consider this as an indicator for \textit{super-human} reliability \cite{Buechel18naacl}.

\begin{table}[t!]
    \centering
    \small
\begin{tabular}{ccc|cccc}
  \rt{\textbf{Gold1}} &   \rt{\textbf{Gold2}} &     \rt{\textbf{Shared}} &        \rt{\textbf{Emo}} &  \rt{\textbf{G1vsG2}} &  \rt{\textbf{G1vsPr}} &  \rt{\textbf{G2vsPr}} \\
\toprule
 \multirow{3}{*}{en1} &  \multirow{3}{*}{en2} &  \multirow{3}{*}{1032} &    V &    \textbf{.953} &      .941 &      .922 \\
  &   &   &    A &    .760 &      \textbf{.761} &      .711 \\
  &   &   &  D &    .794 &      \textbf{.879} &      .782 \\
  \midrule
 \multirow{2}{*}{es1} &  \multirow{2}{*}{es2} &   \multirow{2}{*}{610} &    V &    \textbf{.976} &      .905 &      .912 \\
  &   &    &    A &    \textbf{.758} &      .714 &      .725 \\
  \midrule
 \multirow{2}{*}{es2} &  \multirow{2}{*}{es3} &   \multirow{2}{*}{222} &    V &    \textbf{.976} &      .906 &      .907 \\
  &   &    &    A &    .710 &      \textbf{.724} &      .691 \\
  \midrule
 \multirow{2}{*}{de2} &  \multirow{2}{*}{de3} &   \multirow{2}{*}{498} &    V &    \textbf{.963} &      .806 &      .812 \\
  &   &    &    A &    \textbf{.760} &      .721 &      .663 \\
  \midrule
 \multirow{2}{*}{pl1} &  \multirow{2}{*}{pl2} &  \multirow{2}{*}{445} &    V &    \textbf{.943} &      .838 &      .852 \\
  &   &    &    A &    .725 &      \textbf{.764} &      .643 \\
  \midrule
 \multirow{2}{*}{zh1} &  \multirow{2}{*}{zh2} &   \multirow{2}{*}{140} &    V &    \textbf{.932} &      .918 &      .898 \\
  &   &    &    A &    .482 &      \textbf{.556} &      .455 \\
\bottomrule
\end{tabular}
    \caption{ \label{tab:human_reliability}
    Comparison against human performance. Correlation between two gold standards, \textbf{Gold1} and \textbf{Gold2}, with each other (\textbf{G1vsG2}), as well as with our lexicons \texttt{TargetPred-test} (\textbf{G1vsPr} and \textbf{G2vsPr})  relative to \textbf{Emo}tional variable and  \textbf{Shared} number of words.
    }
\end{table}

As shown in Table \ref{tab:human_reliability}, our lexicons are often competitive with human reliability for Valence (especially for English and Chinese), but outperform human reliability in 4 out of 6 cases for Arousal, and in the single test case for Dominance. There are no cases of overlapping gold standards for BE5.

\section{Methodological Assumptions Revisited}
\label{sec:analysis}
This section investigates patterns in prediction quality \textit{across} languages, validating design decisions of our methodology.

\paragraph{Translation vs.\ Prediction.}
Is it beneficial to predict new ratings for the words in \texttt{TargetMT}  rather than  using them as final lexicon entries straight away? For each \texttt{TargetGold} lexicon (cf.\ Table \ref{tab:gold_lexica}), we intersect its word material with that in \texttt{TargetMT} and \texttt{TargetPred}. Then, we compute the correlation between \texttt{TargetPred} and \texttt{TargetMT} with the gold standard. This analysis was done on the respective \textit{train} sets because using \texttt{TargetMT} rather than \texttt{TargetPred} is only an option for entries known at training time.

Table \ref{tab:translation-vs-prediction} depicts the results of this comparison averaged over all gold lexicons.
As hypothesized, the \texttt{TargetPred} lexicons agree, on average, more with human judgment than the \texttt{TargetMT} lexicons, suggesting that the word emotion model acts as a value-adding post-processor, partly mitigating rating inconsistencies introduced by mere translation of the lexicons. The observation holds for each individual emotion variable with particularly large benefits for Arousal, where the post-processed \texttt{TargetPred} lexicons are on average 14\%-points better compared to the translation-only \texttt{TargetMT} lexicons. This seems to indicate that lexical Arousal is less consistent between translational equivalents compared to other emotional meaning components like Valence and Sadness, which appear to be more robust against translation.

\begin{table}[t!]
\centering
\small
\begin{tabular}{l|p{4mm}p{4mm}p{4mm}p{4mm}p{4mm}p{4mm}p{4mm}p{4mm}}
{} &  \rt{\textbf{Val}} &  \rt{\textbf{Aro}} &  \rt{\textbf{Dom}} &  \rt{\textbf{Joy}} &  \rt{\textbf{Ang}} &  \rt{\textbf{Sad}} &  \rt{\textbf{Fea}} &  \rt{\textbf{Dis}} \\
\toprule
\texttt{Pred} &     .871 &     .652 &       .733 & .767 &   .734 &     .692 &  .728 &     .650 \\
\texttt{MT}   &     .796 &     .515 &       .613 & .699 &   .677 &     .636 &  .654 &     .579 \\
\midrule
Diff       &     .076 &     .137 &       .119 & .068 &   .057 &     .056 &  .074 &     .071 \\
\bottomrule
\end{tabular}
   \caption{ \label{tab:translation-vs-prediction}
   Quality of \texttt{TargetMT} vs.\ \texttt{TargetPred} in terms of average Pearson correlation over all languages and gold standards. $\text{Diff} \coloneqq \text{\texttt{Pred}} - \text{\texttt{MT}}$.
   }
\end{table}

\paragraph{Gold vs.\ Silver Evaluation.}
\label{sec:gold-silver-agreement}

How meaningful is silver evaluation without gold data? We compute the Pearson correlation between gold and silver evaluation results across languages per emotion variable. For languages where we consider multiple datasets during gold evaluation, we first average the gold evaluation results for each emotion variable. As can be seen from Table \ref{tab:silver-gold-corr}, the correlation values range between $r = .91$ for Joy and $r = .27$ for Disgust. This relatively large dispersion is not surprising when we take into account that we correlate very small data series (for Valence and Arousal there are just 12 languages for which both gold and silver evaluation results are available; for BE5 there are only 5 such languages). However, the mean over all correlation values in Table \ref{tab:silver-gold-corr} is $.64$, indicating that there is a relatively strong correlation between both types of evaluation. This suggests that the silver evaluation may be used as a rather reliable proxy of lexicon quality even in the absence of language-specific gold data. 

\begin{table}[h!]
\centering
\small
\vspace*{4pt}
\begin{tabular}{l|rrrrrrrr}
{} & \rt{\textbf{Val}} & \rt{\textbf{Aro}} & \rt{\textbf{Dom}} & \rt{\textbf{Joy}} & \rt{\textbf{Ang}} & \rt{\textbf{Sad}} & \rt{\textbf{Fea}} & \rt{\textbf{Dis}} \\
 \toprule
\#Lg   &   12 &   12 &      8 &  5 &  5 &    5 &  5 &    5 \\
$r$  &    .54 &     .57 &       .52 &  .91 &   .85 &     .57 &  .87 &     .27 \\
\bottomrule
\end{tabular}

\caption{ \label{tab:silver-gold-corr}
Agreement between gold and silver evaluation across languages in Pearson's $r$ relative to the number of applicable languages (``\#Lg'').
}
\end{table}

\section{Conclusion}
\label{sec:conc}

Emotion lexicons are at the core of sentiment analysis, a rapidly flourishing field of NLP. Yet, despite large community efforts, the coverage of existing lexicons is still limited in terms of languages, size, and types of emotion variables. While there are techniques to tackle these three forms of sparsity in isolation, we introduced a methodology which allows us to cope with them simultaneously by jointly combining emotion representation mapping, machine translation, and embedding-based lexicon expansion.

Our study is ``large-scale'' in many respects. We created representationally complex lexicons---comprising 8 distinct emotion variables---for 91 languages with up to 2 million entries each. The evaluation of the generated lexicons featured 26 manually annotated datasets spanning 12 diverse languages. The predicted ratings showed consistently high correlation with human judgment, compared favorably with state-of-the-art monolingual approaches to lexicon expansion and even surpassed human inter-study reliability in some cases.

The sheer number of test sets we used allowed us to validate fundamental methodological assumptions underlying our approach. Firstly, the evaluation procedure, which is integrated into the generation methodology, allows us to reliably estimate the quality of resulting lexicons, \textit{even without target language gold standard}. Secondly, our data suggests that embedding-based word emotion models can be used as a \textit{repair mechanism}, mitigating poor target-language emotion estimates acquired by simple word-to-word translation.

Future work will have to deepen the way we deal with word sense ambiguity by way of exchanging the simplifying type-level approach our current work is based on with a semantically more informed sense-level approach. A promising direction would be to combine a multilingual sense inventory such as \textsc{BabelNet} \cite{Navigli12} with sense embeddings \cite{CamachoCollados18}.

\section*{Acknowledgments}

We would like to thank the anonymous reviewers for their helpful suggestions and comments, and Tinghui Duan, \textsc{Julie Lab}, for  assisting us with the Chinese gold data. This work was partially funded by the German Federal Ministry for Economic Affairs and Energy (funding line ``Big Data in der makroökonomischen Analyse'' [Big data in macroeconomic analysis]; Fachlos 2; GZ 23305/003\#002).

\newpage
\bibliography{literature.bib}
\bibliographystyle{stylesheets/acl2020/acl_natbib}

\clearpage
\appendix
\section{Appendices}
\label{sec:appendix}

\subsection{Data Preparation}
\label{appendix:data}

The exact design of the \texttt{Source} train-dev-test split is as follows: All entries (words plus ratings) from all splits are taken from \citet{Warriner13}. The data was then partitioned based on the overlap with the two precursory versions by \citet{Bradley99anew} (the original \textsc{Anew}) and \citet{Bradley10} (an early extended version of \textsc{Anew} roughly twice as large). \texttt{Source-test} was built by intersecting the lexicon from \citet{Warriner13} with the original \textsc{Anew}. A similar process was applied for \texttt{Source-dev}: we intersected the words from \citet{Warriner13} and \citet{Bradley10} and removed the ones present in \texttt{Source-test}. Lastly, \texttt{Source-train} is made up by all words from \citet{Warriner13} which are neither in \texttt{Source-test} nor in \texttt{Source-dev}. The reason why the ratings in \texttt{Source} are taken exclusively from \citet{Warriner13} is that these are distributed under a more permissive license compared to their precursors.

We removed multi-token entries (e.g., \textit{boa constrictor}) and entries with upper case characters (e.g.,  \textit{Budweiser}) from all data splits of \texttt{Source}, thus restricting the lexicon to single-token, non-proper noun entries to make it more suitable for word embedding-based research. All splits combined have 13,791 entries (train: 11,463, dev: 1,296, test: 1,032), thus removing less than 1\% from the original lexicon.\footnote{The data split is available at: \url{https://github.com/JULIELab/XANEW}}

Regarding the remaining gold standards, the only cases which needed additional preparation or cleansing steps were \texttt{zh1} \citep{Yu16naacl} and \texttt{zh2} \citep{Yao17}. \texttt{zh1} was created and is distributed using traditional Chinese characters, whereas the embedding model by \citet{Grave18lrec} employs simplified ones. Therefore, we converted \texttt{zh1} into simplified characters using \textsc{Google Translate}\footnote{In this case the regular Web application, not the API, was used: \url{https://translate.google.com/}} prior to evaluation.

While manually examining the \texttt{zh2} lexicon, we noticed several cases where the ratings seemed rather counter-intuitive (e.g., seemingly positive words which received very negative ratings). We contacted the authors who confirmed the problem and sent us a corrected version. We did not find any such problems in the second version.
We consulted with a Chinese native speaker for both of these procedures regarding the \texttt{zh1} and \texttt{zh2} lexicons.

\subsection{Model Training and Implementation}
\label{sec:appendix.model}

Training of the MTLFFN model closely followed the procedure specified by \citet{Buechel18naacl}: For each language, the model was trained for roughly 15k iterations (exactly 168 epochs) with a batch size of 128 using the Adam optimizer \cite{Kingma15} with learning rate $10^{-3}$, and $.5$ dropout on the hidden layers and $.2$ on the input layer. As nonlinear activation function we used leaky ReLU with ``leakage'' of $0.01$.

Embedding vectors are the only model input. They have 300 dimensions for every language, independent of their respective training data size \citep{Grave18lrec}. Since the automatic translation of \texttt{Source} is not guaranteed to result in single-word translations, we use the following workaround to derive embedding vectors for multi-token translations: If the translation as a whole cannot be found in the embedding model, the multi-token term gets split up into its constituent parts, using spaces, apostrophes or hyphens as separators. Each substring is looked up in the embedding model, the averaged vector is taken as input. If no substring is recognized, we use the zero vector instead. We also use the zero vector for single-token entries in \texttt{TargetMT} that are missing in the embeddings.

Since \citet{Buechel18naacl} considered only VAD but not BE5 datasets, we conducted a development experiment on the \texttt{TargetMT-dev} sets for all 91 languages where we assessed whether MTL is advantageous for BE5 variables as well, or for a combination of VAD and BE5 variables. We found that MTL improved performance when applied separately among all VAD and BE5 variables. Yet, when jointly learning all eight emotion variables, the results were somewhat inconclusive. Performance \textit{increased} for BE5, but \textit{decreased} for VAD.  Hence, for lexicon creation, we took a cautious approach and trained \textit{two separate models per language}, one for VAD, the other for BE5. An analysis of MTL across VAD and BE5 is left for future work.

The MTLFFN model is implemented in \href{https://pytorch.org/}{\sc PyTorch}, adapting part of the \textsc{Tensorflow} code from \newcite{Buechel18naacl}. The ridge regression baseline model is implemented with \textsc{scikit-learn} \cite{scikit-learn} using default parameters.

\begin{table*}
\tiny
\center
\begin{tabular}{rllrrrrrrrrrr}
\toprule
{\textbf{No.}} & \textbf{ISO} &           \textbf{Full Name} &  \textbf{Size} &  \textbf{Val} &  \textbf{Aro} &  \textbf{Dom} &  \textbf{Joy} &  \textbf{Ang} &  \textbf{Sad} &  \textbf{Fea} &  \textbf{Dis} &  \textbf{Mean} \\
\midrule
1   &   en &             English &    2,000,004 &  .94 &  .76 &  .88 &  .90 &  .91 &  .90 &  .89 &  .89 &   .88 \\
2   &   es &             Spanish &    2,001,183 &  .89 &  .70 &  .80 &  .83 &  .86 &  .85 &  .82 &  .81 &   .82 \\
3   &   it &             Italian &    2,001,137 &  .88 &  .69 &  .81 &  .82 &  .85 &  .84 &  .82 &  .81 &   .81 \\
4   &   de &              German &    2,000,507 &  .89 &  .66 &  .81 &  .82 &  .84 &  .82 &  .80 &  .81 &   .81 \\
5   &   sv &             Swedish &    2,000,980 &  .87 &  .64 &  .80 &  .82 &  .84 &  .82 &  .81 &  .80 &   .80 \\
6   &   pt &          Portuguese &    2,001,078 &  .86 &  .70 &  .78 &  .78 &  .83 &  .81 &  .78 &  .82 &   .79 \\
7   &   id &          Indonesian &    2,002,221 &  .85 &  .67 &  .79 &  .78 &  .82 &  .80 &  .79 &  .77 &   .79 \\
8   &   hu &           Hungarian &    2,000,975 &  .86 &  .67 &  .79 &  .80 &  .82 &  .79 &  .77 &  .79 &   .79 \\
9   &   fr &              French &    2,001,517 &  .85 &  .65 &  .79 &  .78 &  .82 &  .81 &  .78 &  .81 &   .78 \\
10  &   fi &             Finnish &    2,000,841 &  .86 &  .64 &  .79 &  .81 &  .82 &  .78 &  .77 &  .80 &   .78 \\
11  &   ro &            Romanian &    2,001,501 &  .85 &  .65 &  .78 &  .78 &  .82 &  .81 &  .79 &  .78 &   .78 \\
12  &   cs &               Czech &    2,001,203 &  .84 &  .64 &  .77 &  .78 &  .82 &  .80 &  .79 &  .79 &   .78 \\
13  &   pl &              Polish &    2,001,460 &  .85 &  .63 &  .78 &  .80 &  .82 &  .80 &  .78 &  .78 &   .78 \\
14  &   nl &               Dutch &    2,000,721 &  .85 &  .64 &  .78 &  .77 &  .80 &  .79 &  .77 &  .78 &   .77 \\
15  &   no &  Norwegian (Bokmål) &    2,000,876 &  .84 &  .63 &  .77 &  .78 &  .82 &  .78 &  .78 &  .78 &   .77 \\
16  &   tr &             Turkish &    2,002,489 &  .84 &  .62 &  .78 &  .78 &  .80 &  .80 &  .75 &  .77 &   .77 \\
17  &   ru &             Russian &    2,001,317 &  .82 &  .64 &  .75 &  .80 &  .81 &  .77 &  .77 &  .77 &   .77 \\
18  &   el &               Greek &    2,001,704 &  .82 &  .63 &  .76 &  .78 &  .80 &  .78 &  .77 &  .78 &   .77 \\
19  &   uk &           Ukrainian &    2,001,261 &  .83 &  .63 &  .77 &  .78 &  .80 &  .77 &  .76 &  .77 &   .76 \\
20  &   et &            Estonian &    2,001,125 &  .83 &  .59 &  .75 &  .77 &  .81 &  .78 &  .77 &  .78 &   .76 \\
21  &   ca &             Catalan &    2,001,538 &  .84 &  .60 &  .80 &  .77 &  .79 &  .78 &  .76 &  .74 &   .76 \\
22  &   da &              Danish &    2,000,654 &  .84 &  .61 &  .77 &  .78 &  .79 &  .77 &  .73 &  .79 &   .76 \\
23  &   lv &             Latvian &    1,642,923 &  .82 &  .63 &  .75 &  .76 &  .79 &  .78 &  .76 &  .77 &   .76 \\
24  &   lt &          Lithuanian &    2,001,306 &  .83 &  .63 &  .77 &  .75 &  .79 &  .77 &  .75 &  .76 &   .76 \\
25  &   bg &           Bulgarian &    2,001,391 &  .82 &  .60 &  .76 &  .75 &  .77 &  .77 &  .73 &  .76 &   .74 \\
26  &   he &              Hebrew &    2,001,984 &  .80 &  .62 &  .72 &  .76 &  .78 &  .76 &  .74 &  .75 &   .74 \\
27  &   zh &             Chinese &    2,001,799 &  .79 &  .60 &  .75 &  .72 &  .77 &  .75 &  .75 &  .73 &   .73 \\
28  &   mk &          Macedonian &    1,356,402 &  .82 &  .54 &  .75 &  .77 &  .76 &  .73 &  .72 &  .74 &   .73 \\
29  &   af &           Afrikaans &     883,464 &  .80 &  .58 &  .74 &  .76 &  .75 &  .74 &  .71 &  .74 &   .73 \\
30  &   tl &             Tagalog &     716,272 &  .80 &  .56 &  .76 &  .70 &  .77 &  .76 &  .74 &  .72 &   .73 \\
31  &   sk &              Slovak &    2,001,221 &  .80 &  .60 &  .75 &  .74 &  .74 &  .73 &  .71 &  .73 &   .72 \\
32  &   sq &            Albanian &    1,169,697 &  .80 &  .57 &  .73 &  .75 &  .75 &  .75 &  .72 &  .72 &   .72 \\
33  &   az &         Azerbaijani &    2,002,146 &  .81 &  .60 &  .73 &  .74 &  .75 &  .73 &  .70 &  .71 &   .72 \\
34  &   mn &           Mongolian &     608,598 &  .78 &  .57 &  .73 &  .71 &  .78 &  .72 &  .74 &  .74 &   .72 \\
35  &   hy &            Armenian &    2,001,329 &  .80 &  .52 &  .72 &  .75 &  .77 &  .73 &  .71 &  .73 &   .72 \\
36  &   eo &           Esperanto &    2,001,575 &  .77 &  .55 &  .71 &  .72 &  .76 &  .74 &  .73 &  .73 &   .71 \\
37  &   sl &           Slovenian &    1,992,272 &  .81 &  .54 &  .75 &  .74 &  .74 &  .70 &  .70 &  .72 &   .71 \\
38  &   hr &            Croatian &    2,001,570 &  .78 &  .56 &  .71 &  .72 &  .74 &  .71 &  .71 &  .73 &   .71 \\
39  &   gl &            Galician &    1,336,256 &  .78 &  .53 &  .72 &  .72 &  .76 &  .74 &  .71 &  .71 &   .71 \\
40  &   sr &             Serbian &    2,002,395 &  .76 &  .57 &  .71 &  .72 &  .74 &  .70 &  .70 &  .73 &   .70 \\
41  &   ar &              Arabic &    2,003,155 &  .78 &  .53 &  .70 &  .70 &  .75 &  .72 &  .71 &  .74 &   .70 \\
42  &   fa &             Persian &    2,003,533 &  .77 &  .58 &  .70 &  .70 &  .74 &  .73 &  .70 &  .70 &   .70 \\
43  &   ms &               Malay &    1,213,397 &  .75 &  .58 &  .69 &  .69 &  .72 &  .70 &  .65 &  .73 &   .69 \\
44  &   mr &             Marathi &     848,549 &  .74 &  .54 &  .68 &  .70 &  .74 &  .70 &  .69 &  .71 &   .69 \\
45  &   ka &            Georgian &    1,567,232 &  .76 &  .52 &  .72 &  .70 &  .72 &  .71 &  .70 &  .66 &   .69 \\
46  &   ja &            Japanese &    2,003,306 &  .72 &  .58 &  .67 &  .68 &  .71 &  .70 &  .70 &  .68 &   .68 \\
47  &   hi &               Hindi &    1,879,196 &  .76 &  .56 &  .68 &  .69 &  .73 &  .64 &  .65 &  .72 &   .68 \\
48  &   is &           Icelandic &     945,214 &  .76 &  .55 &  .70 &  .68 &  .70 &  .69 &  .68 &  .64 &   .67 \\
49  &   kk &              Kazakh &    1,981,562 &  .72 &  .53 &  .65 &  .67 &  .73 &  .69 &  .67 &  .70 &   .67 \\
50  &   ko &              Korean &    2,002,600 &  .74 &  .57 &  .69 &  .67 &  .67 &  .66 &  .65 &  .69 &   .67 \\
51  &   be &          Belarusian &    1,715,582 &  .73 &  .49 &  .66 &  .68 &  .71 &  .67 &  .67 &  .70 &   .66 \\
52  &   bn &             Bengali &    1,471,709 &  .74 &  .50 &  .67 &  .67 &  .70 &  .67 &  .67 &  .66 &   .66 \\
53  &   kn &             Kannada &    1,747,421 &  .70 &  .47 &  .65 &  .67 &  .71 &  .68 &  .67 &  .68 &   .65 \\
54  &   cy &               Welsh &     502,006 &  .72 &  .51 &  .67 &  .64 &  .69 &  .65 &  .64 &  .66 &   .65 \\
55  &   ur &                Urdu &    1,157,969 &  .69 &  .52 &  .61 &  .63 &  .70 &  .65 &  .64 &  .68 &   .64 \\
56  &   ta &               Tamil &    2,002,514 &  .70 &  .51 &  .66 &  .64 &  .66 &  .66 &  .63 &  .64 &   .64 \\
57  &   eu &              Basque &    1,828,013 &  .70 &  .46 &  .66 &  .64 &  .68 &  .67 &  .64 &  .64 &   .64 \\
58  &   ml &           Malayalam &    2,002,920 &  .67 &  .51 &  .62 &  .63 &  .67 &  .67 &  .62 &  .61 &   .63 \\
59  &   gu &            Gujarati &     557,270 &  .69 &  .46 &  .62 &  .61 &  .67 &  .65 &  .63 &  .64 &   .62 \\
60  &   si &           Sinhalese &     812,356 &  .66 &  .48 &  .59 &  .65 &  .67 &  .62 &  .63 &  .65 &   .62 \\
61  &   te &              Telugu &    1,880,585 &  .69 &  .46 &  .62 &  .60 &  .65 &  .63 &  .61 &  .65 &   .61 \\
62  &   ne &              Nepali &     580,582 &  .68 &  .44 &  .62 &  .63 &  .65 &  .63 &  .61 &  .62 &   .61 \\
63  &   tg &               Tajik &     508,617 &  .67 &  .38 &  .64 &  .57 &  .65 &  .65 &  .60 &  .60 &   .60 \\
64  &   vi &          Vietnamese &    2,008,605 &  .65 &  .47 &  .58 &  .59 &  .65 &  .59 &  .58 &  .62 &   .59 \\
65  &   pa &     Eastern Punjabi &     403,997 &  .67 &  .37 &  .61 &  .59 &  .64 &  .61 &  .58 &  .62 &   .59 \\
66  &   bs &             Bosnian &    1,124,938 &  .63 &  .43 &  .60 &  .57 &  .64 &  .61 &  .61 &  .60 &   .58 \\
67  &   ky &             Kirghiz &     751,902 &  .65 &  .37 &  .61 &  .56 &  .64 &  .62 &  .59 &  .60 &   .58 \\
68  &   ga &               Irish &     321,249 &  .64 &  .47 &  .59 &  .58 &  .61 &  .61 &  .59 &  .55 &   .58 \\
69  &   fy &        West Frisian &     530,054 &  .61 &  .43 &  .54 &  .53 &  .60 &  .59 &  .55 &  .58 &   .56 \\
70  &   uz &               Uzbek &     833,860 &  .60 &  .38 &  .55 &  .56 &  .57 &  .56 &  .54 &  .53 &   .53 \\
71  &   sw &             Swahili &     391,312 &  .59 &  .34 &  .57 &  .52 &  .59 &  .58 &  .57 &  .51 &   .53 \\
72  &   jv &            Javanese &     518,634 &  .58 &  .45 &  .53 &  .53 &  .56 &  .58 &  .54 &  .49 &   .53 \\
73  &   ps &              Pashto &     300,927 &  .58 &  .40 &  .56 &  .52 &  .55 &  .54 &  .55 &  .49 &   .53 \\
74  &   am &             Amharic &     308,109 &  .56 &  .31 &  .52 &  .48 &  .53 &  .54 &  .52 &  .47 &   .49 \\
75  &   lb &       Luxembourgish &     642,504 &  .53 &  .37 &  .47 &  .45 &  .55 &  .52 &  .50 &  .51 &   .49 \\
76  &   su &           Sundanese &     327,533 &  .54 &  .36 &  .47 &  .45 &  .53 &  .52 &  .48 &  .52 &   .48 \\
77  &   th &                Thai &    2,006,540 &  .51 &  .38 &  .45 &  .50 &  .49 &  .46 &  .45 &  .49 &   .47 \\
78  &   km &               Khmer &     247,498 &  .51 &  .39 &  .44 &  .49 &  .51 &  .44 &  .45 &  .48 &   .46 \\
79  &   sd &              Sindhi &     139,063 &  .47 &  .35 &  .39 &  .41 &  .50 &  .49 &  .50 &  .46 &   .45 \\
80  &   yi &             Yiddish &     205,727 &  .49 &  .34 &  .40 &  .43 &  .50 &  .47 &  .45 &  .44 &   .44 \\
81  &   my &             Burmese &     339,628 &  .49 &  .36 &  .42 &  .43 &  .49 &  .45 &  .46 &  .43 &   .44 \\
82  &   la &               Latin &    1,088,139 &  .47 &  .33 &  .40 &  .39 &  .47 &  .46 &  .43 &  .44 &   .42 \\
83  &   mt &             Maltese &     204,630 &  .47 &  .32 &  .44 &  .38 &  .43 &  .40 &  .39 &  .38 &   .40 \\
84  &   gd &     Scottish Gaelic &     150,694 &  .45 &  .36 &  .39 &  .40 &  .36 &  .36 &  .35 &  .33 &   .38 \\
85  &   so &              Somali &     177,405 &  .40 &  .22 &  .35 &  .36 &  .44 &  .41 &  .41 &  .38 &   .37 \\
86  &   mg &            Malagasy &     415,050 &  .40 &  .32 &  .36 &  .34 &  .41 &  .37 &  .36 &  .36 &   .37 \\
87  &   ht &             Haitian &     118,302 &  .39 &  .22 &  .33 &  .30 &  .42 &  .42 &  .37 &  .38 &   .35 \\
88  &   ku &  Kurdish (Kurmanji) &     395,645 &  .37 &  .22 &  .33 &  .33 &  .34 &  .33 &  .31 &  .35 &   .32 \\
89  &  ceb &             Cebuano &    2,006,001 &  .34 &  .22 &  .29 &  .34 &  .36 &  .32 &  .33 &  .34 &   .32 \\
90  &   co &            Corsican &     108,035 &  .29 &  .24 &  .27 &  .27 &  .32 &  .30 &  .29 &  .30 &   .29 \\
91  &   yo &              Yoruba &     156,764 &  .24 &  .08 &  .19 &  .18 &  .24 &  .21 &  .21 &  .26 &   .20 \\
\bottomrule
\end{tabular}

\caption{\label{tab:generated-lexica}
Overview of generated emotion lexicons with silver evaluation results; sorted by \textbf{Mean} performance over the eight emotional variables.
}
\end{table*}

\end{document}